\documentclass[sigconf]{acmart}
\usepackage{color,soul}
\usepackage{booktabs}
\usepackage{tabularx}
\usepackage{enumitem}
\usepackage{graphicx}
\usepackage{amsmath}
\usepackage{multirow}
\usepackage{array}
\usepackage{wrapfig}
\usepackage{subcaption}
\usepackage{gensymb}
\usepackage{lipsum}
\usepackage{fontspec}
\usepackage{booktabs}
\usepackage{polyglossia}
\usepackage{array}
\usepackage{float}
\setmainlanguage{english}
\setotherlanguage{hindi}
\usepackage{svg}
\usepackage{graphicx}
\usepackage{tcolorbox}
\usepackage{hyperref}
\usepackage{tikz}
\usetikzlibrary{positioning, arrows.meta}

\AtBeginDocument{%
  }

\setcopyright{acmlicensed}
\copyrightyear{2024}
\acmYear{2024}
\acmDOI{XXXXXXX.XXXXXXX}
\acmPrice{15.00}
\acmISBN{978-1-4503-XXXX-X/26/08}
\title{DohaScript: A Large-Scale Multi-Writer Dataset for Continuous Handwritten Hindi Text}

\author{Kunwar Arpit Singh}
\authornote{Both authors contributed equally to this research.}
\orcid{1234-5678-9012-3456} 
\email{kunwar22@iiserb.ac.in}
\affiliation{%
  \institution{IISER Bhopal}
  \city{Bhopal}
  \country{India}
}

\author{Ankush Prakash}
\authornotemark[1]
\email{ankushs22@iiserb.ac.in}
\affiliation{%
  \institution{IISER Bhopal}
  \city{Bhopal}
  \country{India}
}

\author{Haroon R. Lone}
\email{haroon@iiserb.ac.in}
\affiliation{%
  \institution{IISER Bhopal}
  \city{Bhopal}
  \country{India}
}
\begin{document}

\begin{abstract}
Despite hundreds of millions of native speakers, handwritten Devanagari, the script used for Hindi, remains significantly underrepresented in public benchmarks, which predominantly focus on isolated characters and lack writer diversity. We introduce \textit{DohaScript}, a parallel multi-writer dataset of handwritten Hindi text collected from 531 contributors. By having every writer transcribe the same six traditional Hindi couplets (\textit{dohas}), \textit{DohaScript} enables systematic analysis of writer-specific stylistic variation while holding linguistic content constant. The dataset features non-identifiable demographic metadata, objective quality curation based on image resolution and sharpness, and page-level layout difficulty annotations for stratified evaluation. Baseline evaluations reveal that continuous handwritten Devanagari recognition remains challenging even for contemporary OCR systems. A word-level generative baseline further demonstrates the utility of the dataset for writer-conditioned handwriting generation under held-out writer conditioning, while distributional evaluation provides evidence of similarity between real and generated handwriting. A representative subset of the dataset is provided in the \textbf{supplementary material for review}; the complete dataset and accompanying code will be publicly released upon acceptance. \textit{DohaScript} aims to provide a controlled, quality-validated benchmark for advancing continuous handwritten Devanagari research in low-resource settings.
\end{abstract}

\maketitle

\section{Introduction}
\label{sec:intro}
Communication has played a central role in human evolution, and handwriting remains one of its most enduring manifestations, continuing to be extensively used in education, administration, and everyday documentation despite the widespread adoption of digital technologies~\cite{James2017,WileyRapp2021}. Handwriting is also a personal and evolving form of expression that changes under the influence of age, gender, and socio-cultural context, and this inherent variability makes it a challenging object for systematic and computational analysis~\cite{Sharma2022,plamondon2002online,Xing2016DeepWriter}, motivating growing interest in optical character recognition (OCR), handwritten text recognition (HTR), handwriting generation, and writer identification ~\cite{shi2016end, afkari2025comprehensive, dai2023disentangling, Xing2016DeepWriter}. However, most existing work focuses on Latin-script languages, leaving many widely used writing systems insufficiently studied.

\begin{figure}[!t]
    \centering
    \includegraphics[scale=0.14]{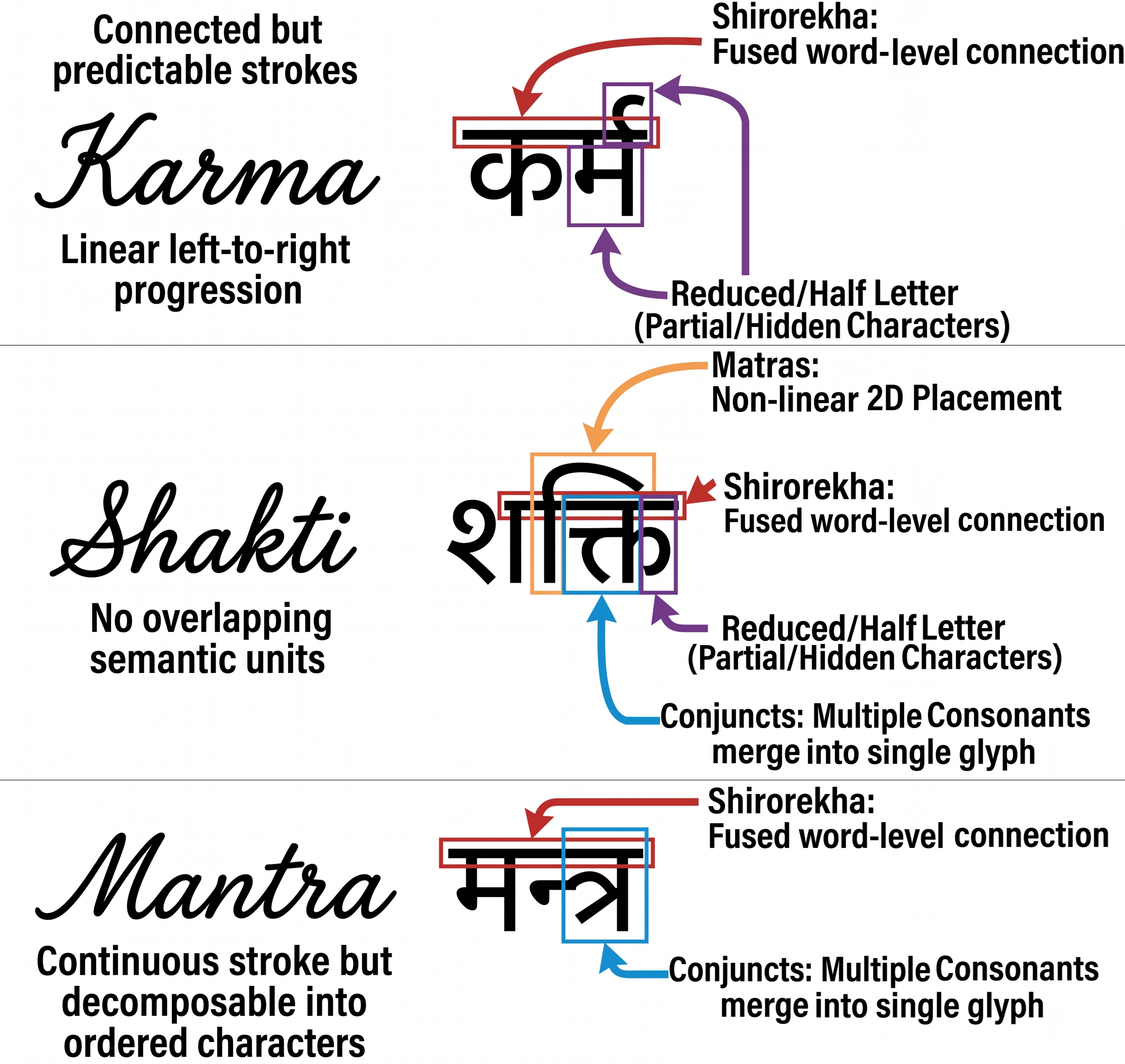}
    \caption{Structural differences between cursive Latin handwriting and handwritten Devanagari script, highlighting shirorekha connectivity, matras, reduced characters, and conjunct glyphs.}
    \label{fig:devanagari_structure}
\end{figure}

This imbalance is particularly evident for Indic languages such as Hindi, Marathi, Nepali, Sanskrit, and Konkani, which are written in the Devanagari script and collectively serve hundreds of millions of speakers~\cite{Shakya2025,Arora2025}. Unlike Latin scripts, Devanagari's structural properties, a continuous shirorekha connecting characters, two-dimensional matra placement, and non-linear conjunct glyphs, make OCR assumptions developed for Latin scripts to transfer poorly (see Figure~\ref{fig:devanagari_structure}). These challenges are further amplified by substantial inter-writer variability in stroke width, slant, spacing, and character formation, making robust representation learning particularly difficult~\cite{PalChaudhuri2004}.

Progress in handwritten Devanagari text analysis is therefore limited not only by the script's structural complexity, but also by the lack of controlled writer-attributed benchmarks for systematic analysis of handwriting variation independent of linguistic content~\cite{Senthil2020}.

Data scarcity poses a major obstacle to modern data-driven approaches, particularly deep learning models that require diverse datasets for effective generalization ~\cite{Sun2017}. Limited training data restricts the ability of models to learn robust representations and leads to poor performance under variations in stroke width, slant, spacing, and writing style. These challenges are further amplified in Devanagari due to its large character inventory, numerous conjunct forms, and positional diacritics ~\cite{PalChaudhuri2004}, so capturing this structural and stylistic diversity reliably requires substantially larger and more heterogeneous datasets.

Despite Hindi being the fourth most widely spoken language worldwide, with over 400 million native speakers and more than 600 million users of Devanagari-based languages~\cite{CensusIndia2011LanguageAtlas,Ethnologue2025}, progress in handwriting recognition and generation remains slow. The lack of standardized public datasets has fragmented research efforts, forcing reliance on private, institution-specific collections and hindering fair comparison across methods~\cite{dutta2018iiit-hw-dev,samanta2024aio-hb}. 
Existing public Devanagari handwriting datasets emphasize lexical diversity or isolated character- and word-level recognition rather than writer-attributed, page-level collection (Table~\ref{tab:dataset_comparison}), leaving no resource suited to controlled analysis of writer-specific variation.

\begin{table*}[]
\centering
\caption{Quantitative comparison of existing Indic handwriting datasets with the proposed dataset.}
\label{tab:dataset_comparison}
\resizebox{\linewidth}{!}{
\begin{tabular}{lccccccc}
\hline
\textbf{Dataset} &
\textbf{Script(s)} &
\textbf{Page-level Samples (\# Pages)} &
\textbf{\# Writers} &
\textbf{Text Type} &
\textbf{Demographics} \\
\hline

IIIT-HW-Words \cite{GongidiJawahar2021IIITIndicHWWords} &
Indic (multi) &
-- (word-only dataset) &
135 &
Isolated words &
No \\

IIIT-HW-Dev \cite{dutta2018iiit-hw-dev} &
Hindi (Devanagari) &
-- (word-only dataset) &
12 &
Continuous words &
Limited \\

Numerals/Vowels Dataset \cite{Prashanth2022DevanagariDataset} &
Devanagari &
-- (character-only dataset) &
$\sim$3800 &
Isolated characters &
Yes\\

CHIPS (PLATTER) \cite{Kasuba2025PLATTER} &
Indic (multi) &
25,290 pages &
-- &
Continuous &
No \\

Parimal Hindi \cite{Dey2025LowResourceHTR} &
Hindi (Devanagari) &
500 pages &
10 &
Continuous paragraphs &
Yes \\

AnciDev \cite{sharma2025ancidev} &
Ancient Devanagari &
500 pages &
-- &
Continuous lines &
No \\

\textbf{DohaScript (ours)} &
\textbf{Hindi (Devanagari)} &
\textbf{531 pages} &
\textbf{531} &
\textbf{Continuous couplets} &
\textbf{Yes} \\

\hline
\end{tabular}
}
\end{table*}

To address these limitations, we introduce \emph{DohaScript}, a multi-writer dataset of handwritten Hindi text collected from 531 unique contributors, making it a real, writer-attributed, page-level Devanagari handwriting resource. In contrast to existing datasets that emphasize lexical diversity, \emph{DohaScript} holds lexical content constant across all writers: each contributor transcribed the same six traditional Hindi \textit{dohas} (couplets), yielding 531 samples of a fixed 89-word corpus. This \emph{parallel stylistic corpus} design isolates handwriting style from linguistic content by construction, enabling controlled evaluation of writer-specific representation learning, OCR robustness, and content-style disentanglement in handwriting synthesis without confounding language-model priors.

\emph{DohaScript} offers four key contributions: (1) \textbf{Dataset design:} A parallel stylistic corpus of 531 page-level real handwriting samples with non-identifiable demographic metadata (age, gender, state).  (2) \textbf{Baseline OCR experiments:} OCR benchmarking across Tesseract, TrOCR, and Sarvam AI highlights the challenges of continuous handwritten Devanagari recognition, even for the strongest evaluated system.  (3) \textbf{Handwriting generation baseline:} A word-level GANWriting synthesis baseline presented as a proof-of-utility demonstration rather than a SOTA generator, showing that DohaScript's shared-lexicon design supports learning writer conditioned representations that render different words drawn from the same shared, fully-seen vocabulary while conditioned on a held-out writer's style.  (4) \textbf{Data validation:} An automated Laplacian-variance and CNN-based curation pipeline, complemented by page-level segmentation difficulty labels, with both components independently validated by human annotators through a two rater study. The dataset is not intended to replace lexically diverse HTR corpora, but to complement them through controlled evaluation of stylistic variability.

\section{Related Datasets}
\label{sec:related}

Existing handwritten text recognition datasets span a wide range of scripts, annotation granularities, and collection methodologies, and for Devanagari and other Indic scripts they fall broadly into three categories: character- and word-level benchmarks, which dominate the landscape; page-level datasets, which are either synthetic or extremely limited in writer diversity; and writer-attributed corpora, which are nearly absent for Devanagari (Table~\ref{tab:dataset_comparison}). At the character and word level, IIIT-HW-Words provides a widely used benchmark of handwritten word images across multiple Indic scripts~\cite{GongidiJawahar2021IIITIndicHWWords}, CALAM offers isolated character-level resources for Indic and Perso-Arabic scripts~\cite{NehraNainAhmed2016DevnagariDatabase}, and IIIT-HW-Dev provides continuous Devanagari words but from only 12 writers~\cite{dutta2018iiit-hw-dev}. None of these datasets capture the shirorekha-induced inter-word connectivity, baseline drift, and contextual dependencies of realistic continuous handwriting~\cite{AlKendi2024HTRSurvey}. At the page level, CHIPS (PLATTER) provides 25,290 synthetically generated pages built by aggregating word-level samples from existing datasets~\cite{Kasuba2025PLATTER}, which enables layout-aware recognition at scale but lacks real writer-attributed stylistic variation by construction, while Parimal Hindi offers 500 real page-level samples in continuous paragraph form from only 10 writers~\cite{Dey2025LowResourceHTR} which is insufficient for studying writer generalization. Our OCR experiments show that synthetic handwriting substantially overestimates real-world accuracy (Section~\ref{sec:ocr_baseline}), directly motivating authentic, writer-attributed page-level data.

Writer-centric directions (e.g., writer-conditioned generation) remain under-supported in existing Devanagari datasets~\cite{Sharma2020IndicScriptsSurvey}, which requires two properties: multiple writers producing identical or overlapping content to separate writing style from text, and genuine handwriting preserving writer-specific stroke characteristics. Existing Indic resources generally lack one or both, limiting controlled writer-centric evaluation. Generative synthesis models additionally require repeated lexical content across many writers ~\cite{Xing2016DeepWriter,dai2023disentangling}, a condition DohaScript satisfies by design, demonstrated with a GANWriting baseline (Section~\ref{sec:ganwriting_baseline}) that exploits this fixed lexical structure to learn writer-specific Devanagari style representations. Taken together, existing resources leave a clear gap: no publicly available dataset provides real, page-level continuous handwriting from a large, writer-attributed, demographically diverse population under controlled lexical conditions which is the precise combination required for studying writer-specific representation learning, OCR robustness, and content-style disentanglement in Devanagari handwriting synthesis. DohaScript is designed to fill this gap.

\section{Dataset Construction}
\label{sec:data_collection}
The DohaScript dataset was designed to control lexical content while preserving natural variation in writers and handwriting styles. The following subsections describe the corpus design, participant recruitment and data acquisition, and the construction of a controlled synthetic reference set.

\subsection{Corpus Design}
We selected six traditional Hindi \textit{dohas} composed by Kabir Das, Rahim Das, and Tulsidas. Their cultural familiarity and widespread use in school curricula make them accessible to a broad participant demographic. More importantly, the corpus was designed to provide broad coverage of Devanagari orthographic units within a compact, naturally flowing text. The six \textit{dohas} (shown below) collectively cover all five consonant articulation categories, the major vowel classes and diacritics, three common conjunct characters, and both Devanagari punctuation symbols. The English meanings of the six \textit{dohas} are provided in Section 1 in Supplementary Material.

\begin{figure*}[t]
    \centering

    \begin{tcolorbox}[
        colframe=black,
        colback=white,
        boxrule=0.8pt,
        arc=0pt,
        boxsep=0pt,      
        left=0pt,
        right=0pt,
        top=0pt,
        bottom=0pt,
        width=0.93\linewidth,
        center
    ]

    \centering
    \includegraphics[width=0.3\linewidth]{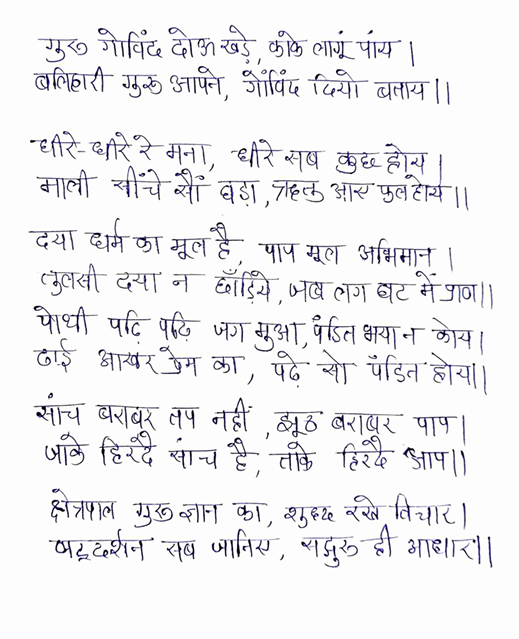}
    \hfill
    \vrule width 0.8pt
    \hfill
    \includegraphics[width=0.3\linewidth]{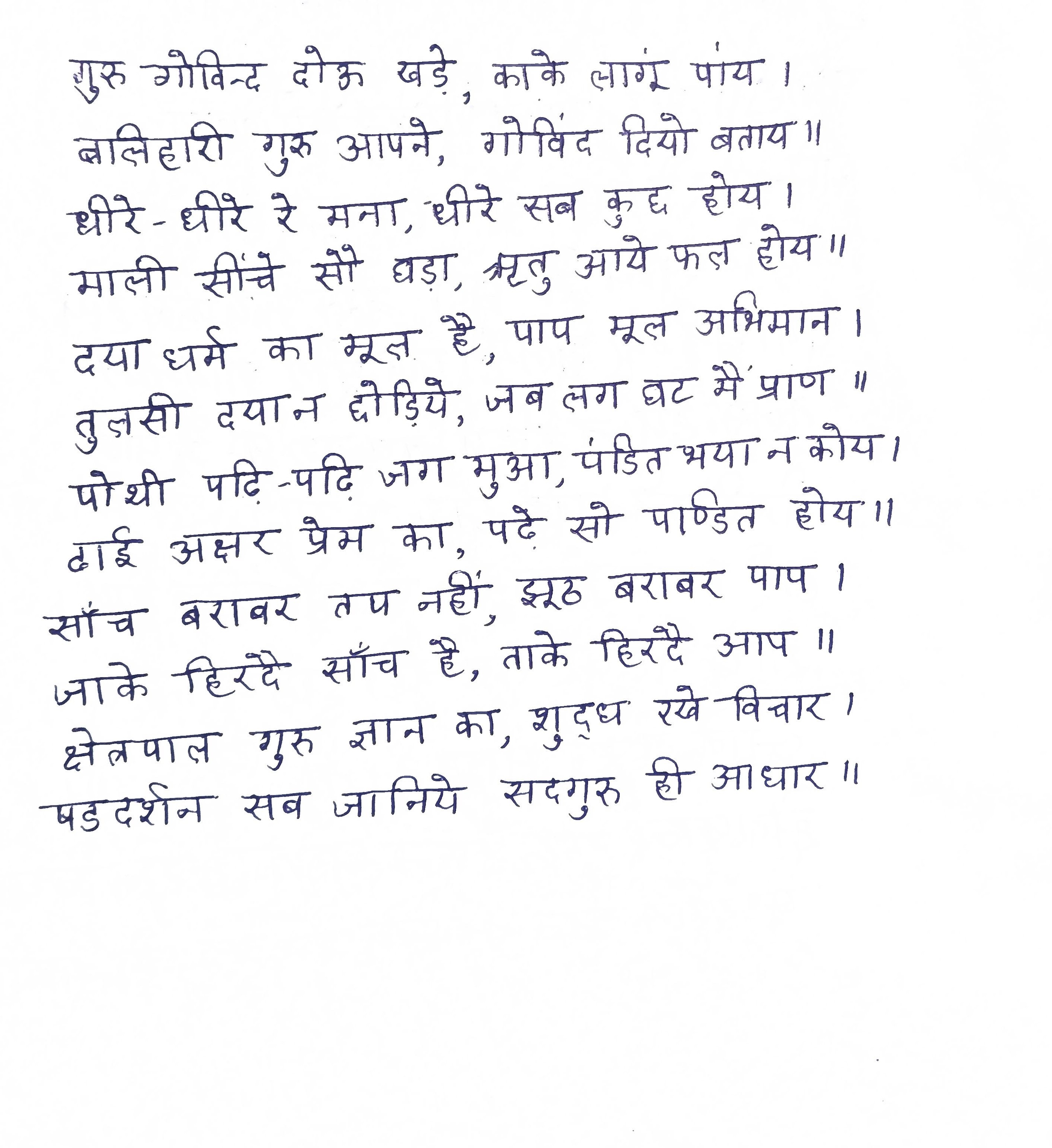}
    \hfill
    \vrule width 0.8pt
    \hfill
    \includegraphics[width=0.3\linewidth]{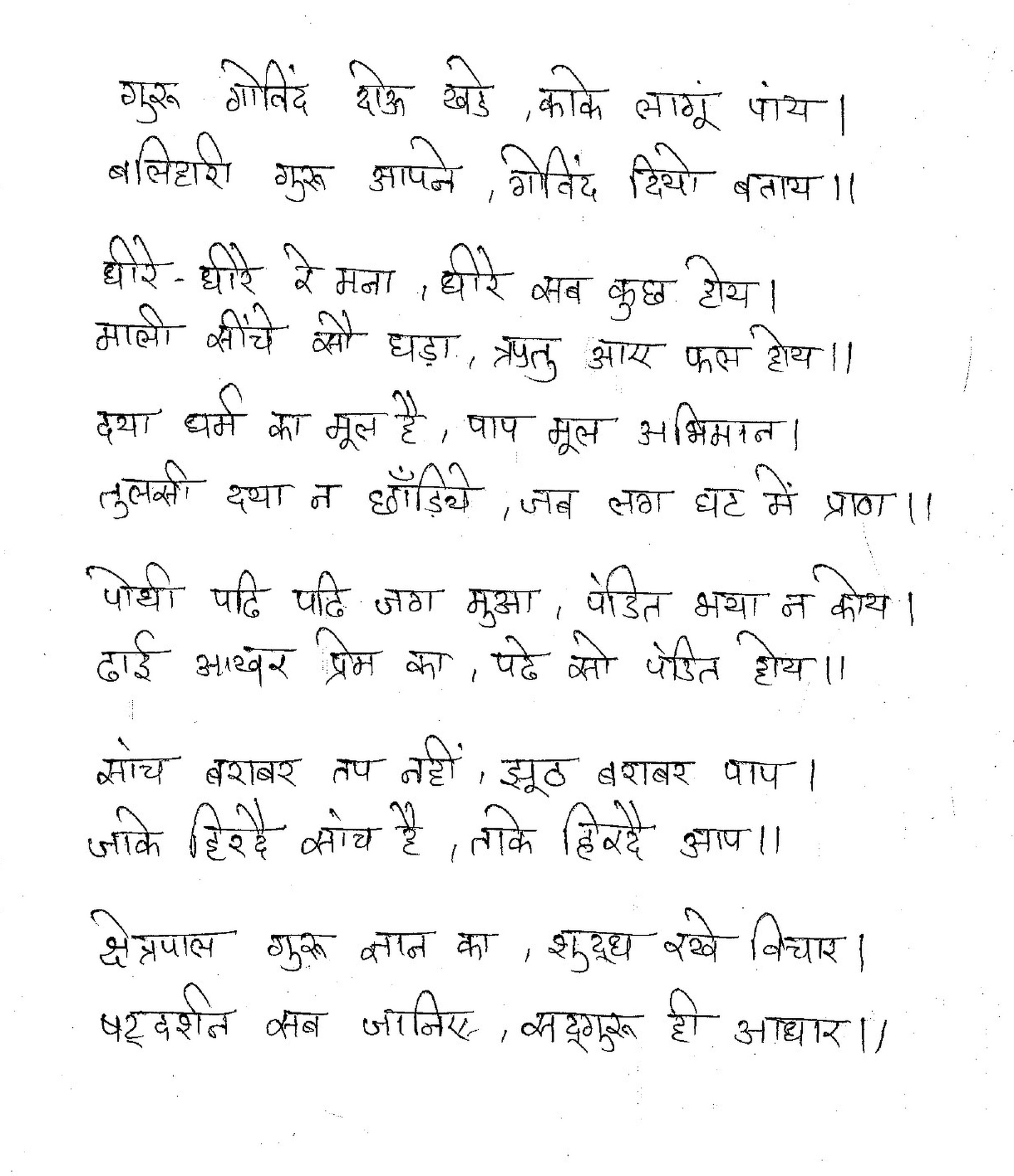}

    \end{tcolorbox}

    \caption{Sample images from the dataset. Each segment corresponds to one handwritten page instance.}
    \label{fig:sample_images}
\end{figure*}

\includegraphics[width=0.95\linewidth]{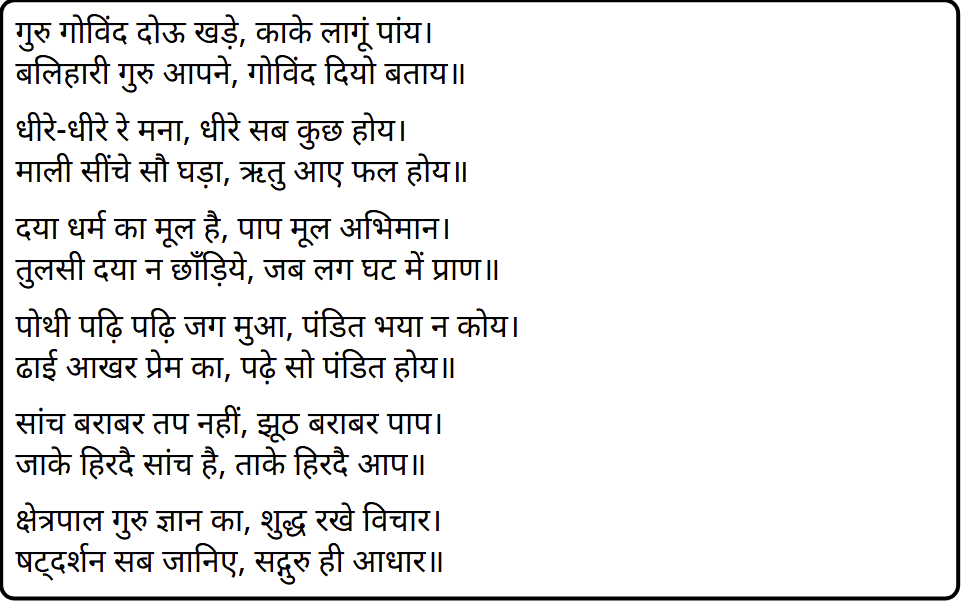}

\hspace*{-2mm}\includegraphics[width=\linewidth]{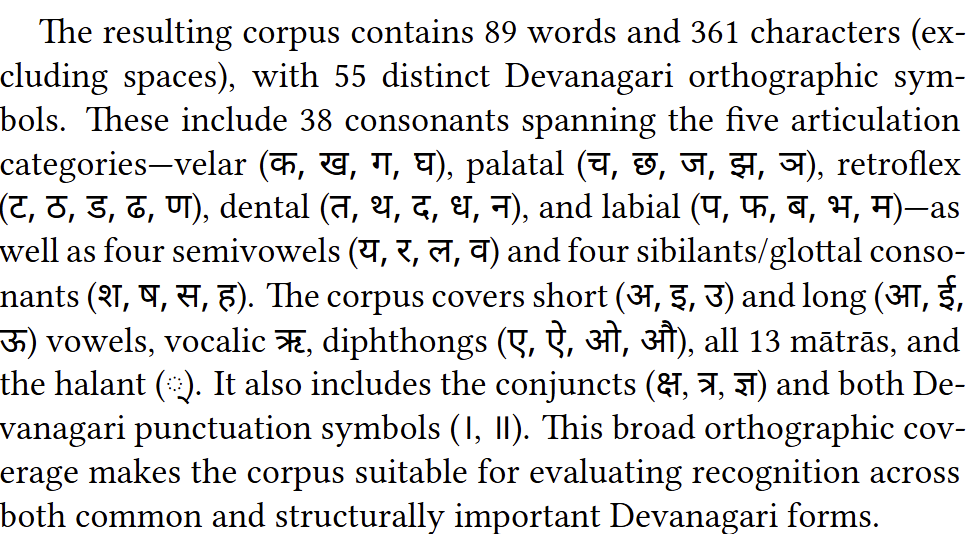}

\subsection{Data Collection}
The dataset was collected from 531 participants, each contributing one handwritten A4 page containing the fixed \textit{doha} corpus. Pages were collected through two channels: some contributors submission via a Google Form as scanned images or mobile photographs, while other collected manually as physical pages and digitized using a scanner. For consistency, all pages were standardized to $2480\times3508$ pixels at 300 DPI.

Participants were recruited through the authors' host institution from five educational institutions across India: one national-level institute in Madhya Pradesh, two state-level colleges in Bihar, and two elementary schools in Uttar Pradesh. Recruitment was conducted through direct contact with teachers and students, with entirely voluntary participation. No restrictions were imposed on handwriting proficiency, allowing natural variation in handwriting styles.

Only non-identifiable demographic information was collected, including age, gender, and city. The final cohort comprises 531 participants (135 female and 396 male), with contributors from 24 states and union territories across India. The participant pool was primarily recruited from educational institutions and is consequently skewed toward younger writers. It also exhibits a male-to-female ratio of approximately 3:1. These characteristics reflect the recruitment process and voluntary nature of participation and should be considered when interpreting demographic or subgroup analyses. The age distribution and state-wise participation are provided in Sections  2 and 3 in Supplementary Material.

The dataset contains no names, signatures, or free-form personal text and all samples are restricted to the same fixed couplets, thereby reducing privacy and content-leakage risks. Although the dataset includes participants from a broad range of regions, geographic coverage is uneven, with a substantial proportion originating from Uttar Pradesh and Bihar. Both states are predominantly Hindi-speaking regions where Devanagari is widely used, providing strong representation of native Devanagari writers while retaining geographic diversity across India. Representative samples of DohaScript are shown in Figure~\ref{fig:sample_images}.

\noindent{\bf Ethical Considerations:} The study was approved by the Institutional Ethics Committee of the authors' host institution. Participants were informed of the study's purpose and dataset use, with consent obtained through teachers for minors. No personally identifiable information was collected.
 
\subsection{Synthetic Reference Dataset}
\label{sec:synthetic}

To determine whether OCR errors on DohaScript stem from inherent limitations of recognition systems or the intrinsic difficulty of unconstrained human handwriting, we constructed a controlled synthetic reference dataset. This consisted of 60 page images generated using OpenAI GPT Image 2, a state-of-the-art multimodal generative model, through prompt-based handwriting synthesis, preserving the identical six-\textit{doha} corpus used in DohaScript.

Generation prompts enforced a scanned-document appearance on a plain white A4 background while explicitly varying writer characteristics, including stroke thickness, slant, spacing, baseline stability, pen pressure, and character morphology, to span a broad range of handwriting styles and proficiency levels. This produces a controlled counterfactual in which the same text as DohaScript is generated with reduced writer variability and cleaner acquisition conditions.

The comparison between OCR performance on this synthetic set and on the authentic DohaScript pages (Section~\ref{sec:ocr_baseline}) provides direct empirical evidence that performance gaps reflect genuine handwriting complexity rather than system limitations, further motivating the need for authentic, writer-attributed benchmark data.

\section{Baseline OCR Experiments}
\label{sec:ocr_baseline}
This section addresses two research questions. \textbf{RQ1:} Do existing OCR systems, both traditional pipelines and modern neural models, perform adequately on continuous handwritten Devanagari text, or does DohaScript reveal a performance gap that motivates script-specific development? \textbf{RQ2:} Do observed errors reflect genuine handwriting complexity, or are they artifacts of system limitations that would disappear on cleaner data? We answer RQ1 through a multi-system benchmark on the full DohaScript collection ($N=531$), and RQ2 through a controlled comparison against a synthetic Devanagari reference dataset. All experiments use the fixed six-\textit{doha} corpus as ground truth, enabling standardized comparison via Character Error Rate (CER) and Word Error Rate (WER). CER and WER  are computed as the edit-distance ratio at character and word level respectively (definitions are provided in Section 4 in Supplementary Material).

\subsection{Experimental Setup}

We benchmark OCR approaches spanning traditional pipelines and neural
sequence-to-sequence models. Each model
is evaluated under its intended operating conditions rather than using a
common preprocessing pipeline.

\textbf {Tesseract.} We evaluate four configurations of Tesseract~\cite{smith2007overview}:
\texttt{default} (Hindi language pack with automatic page segmentation),
\texttt{OEM 3, PSM 6}, \texttt{PSM 4}, and bilingual
\texttt{hin+eng} recognition.

\textbf {TrOCR.} We evaluate base-handwritten TrOCR~\cite{li2023trocr},
a Transformer based encoder--decoder model pretrained and fine-tuned on
English handwriting. We apply this checkpoint directly to Devanagari
handwriting without Devanagari-specific pretraining or fine-tuning,
constituting a zero-shot cross-script evaluation.

\textbf {Sarvam AI.} We additionally evaluate Sarvam Document Intelligence~\cite{sarvam2026documentai}, a contemporary OCR
system designed for Indic scripts. Images are submitted directly to the
system without external preprocessing. 

Detailed preprocessing, configuration, decoding, and output normalization
procedures for all systems are provided in Section~6 of the Supplementary
Material.

\subsection{Results}

\begin{table}[t]
\centering
\caption{OCR benchmark results on \emph{DohaScript} ($N{=}531$). Lower is better for both CER and WER.}
\label{tab:ocr_results}
\begin{tabular}{lcc}
\hline
\textbf{Method} & \textbf{CER} $\downarrow$ & \textbf{WER} $\downarrow$ \\
\hline
Sarvam AI (\texttt{hi-IN}) & \textbf{0.1379} & \textbf{0.2719} \\
Tesseract PSM 4 & 0.5940 & 1.0383 \\
Tesseract OEM 3, PSM 6 & 0.7161 & 1.2790 \\
Tesseract Hindi + English & 0.8137 & 1.2508 \\
TrOCR (zero-shot) & 0.9950 & 1.0000 \\
Tesseract Default & 1.0000 & 1.0000 \\
\hline
\end{tabular}
\end{table}

Table~\ref{tab:ocr_results} summarizes the benchmark results. All evaluated Tesseract configurations exhibit high recognition error rates (CER values ranging from 0.594 to 1.000).  The best-performing configuration, \texttt{PSM,4}, achieves a CER of 0.594, but still produces substantial word-level errors, reflected in a WER exceeding 1.0. These results indicate significant difficulty in segmenting and recognizing continuous handwritten Devanagari text under unconstrained writing styles.

TrOCR performs poorly in the zero-shot setting, achieving a CER of 0.995 and frequently producing invalid outputs consisting of Latin characters or numerals. This suggests limited cross-script transfer from English handwriting pretraining to handwritten Devanagari recognition.

Among all evaluated models, Sarvam AI achieves the strongest performance with a CER of 0.138 and WER of 0.272, substantially outperforming both traditional OCR pipelines and zero-shot transformer based recognition. Relative to the best Tesseract configuration, Sarvam AI reduces CER by approximately $4.3\times$ and WER by approximately $3.8\times$. Despite this improvement, the remaining error rates indicate that handwritten Devanagari OCR in realistic settings remains a challenging open problem.

Across all evaluated systems, no model achieves adequate performance for practical handwritten Devanagari recognition. Latin-centric models (Tesseract, TrOCR) effectively fail, while the best Indic-specific system (Sarvam AI), despite achieving CER~$=0.138$, still produces recognition errors on more than one in four words (WER~$=0.272$), demonstrating that continuous handwritten Devanagari remains a non-trivial open problem even for models designed for Indic scripts. This gap motivates both the development of Devanagari-specific OCR models and, critically, the availability of authentic benchmark data to evaluate them, which DohaScript provides.

\subsection{Evaluation on Synthetic Handwritten Data}

To assess whether the observed OCR errors arise primarily from the challenges of unconstrained handwriting rather than limitations of the OCR system itself, we additionally evaluated the best-performing model (Sarvam AI) on a synthetic handwritten Devanagari dataset consisting of 60 generated page images (described in Section~\ref{sec:synthetic}). The synthetic images preserve the same textual content as the six-\textit{doha} corpus while exhibiting substantially cleaner handwriting and reduced writer variability.


Sarvam AI achieves CER~$=0.0649$ and WER~$=0.1424$ on synthetic handwriting, compared to CER~$=0.1379$ and WER~$=0.2719$ on authentic DohaScript pages. Error rates on authentic handwriting are therefore approximately 2.1$\times$ higher in CER and 1.9$\times$ higher in WER, directly answering RQ2. This gap is not attributable to system limitations, as the model performs substantially better on cleaner synthetic inputs, but rather to the genuine complexity of unconstrained human handwriting. The relatively high standard deviations on the synthetic set (CER $\sigma$ $=0.048$, WER $\sigma$\ $=0.114$) further indicate that even synthetic handwriting presents non-trivial variability. Benchmarks based solely on synthetic data would therefore substantially overestimate real-world OCR accuracy, confirming that authentic, writer-attributed corpora such as DohaScript are necessary for meaningful progress measurement.

\section{Handwriting Generation Baseline}
\label{sec:ganwriting_baseline}



\begin{figure}[t]
    \centering

    \setlength{\tabcolsep}{2pt}

    \begin{tabular}{cccc}
        \includegraphics[width=0.23\linewidth]{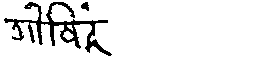} &
        \includegraphics[width=0.23\linewidth]{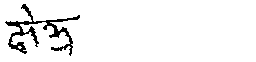} &
        \includegraphics[width=0.23\linewidth]{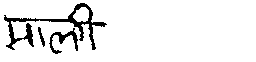} &
        \includegraphics[width=0.23\linewidth]{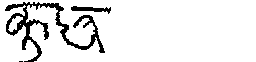} \\

        \includegraphics[width=0.23\linewidth]{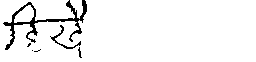} &
        \includegraphics[width=0.23\linewidth]{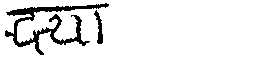} &
        \includegraphics[width=0.23\linewidth]{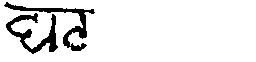} &
        \includegraphics[width=0.23\linewidth]{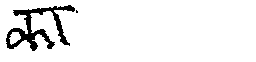} \\

        \includegraphics[width=0.23\linewidth]{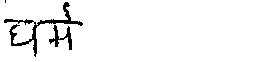} &
        \includegraphics[width=0.23\linewidth]{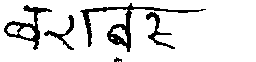} &
        \includegraphics[width=0.23\linewidth]{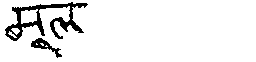} &
        \includegraphics[width=0.23\linewidth]{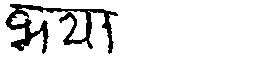} \\

        \includegraphics[width=0.23\linewidth]{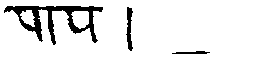} &
        \includegraphics[width=0.23\linewidth]{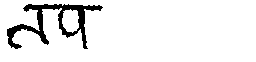} &
        \includegraphics[width=0.23\linewidth]{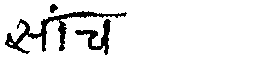} &
        \includegraphics[width=0.23\linewidth]{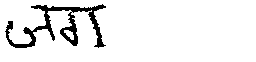}
    \end{tabular}

    \caption{Examples of Hindi handwritten word images generated using the adapted
GANWriting model on the DohaScript dataset, illustrating variation in
stroke thickness, slant, and \emph{shirorekha} appearance.}
    \label{fig:ganwriting_results}
\end{figure}

To demonstrate the utility of \emph{DohaScript} for writer-conditioned
handwriting generation, we adapt GANWriting~\cite{kang2020ganwriting}
and train it on word-level samples. The fixed lexical corpus enables
writer conditioning while holding textual content constant; model,
training, and word-level preprocessing details are provided in the
Section 8 in Supplementary Material. Qualitative examples of generated
handwritten Hindi words are shown in Figure~\ref{fig:ganwriting_results},
illustrating variation in handwriting appearance across generated samples.

\subsection{Quantitative Results}
\label{sec:ganwriting_quant}
Training reduced CER from 100\% at initialization to 15.58–32.31\% for independently sampled words at epoch 50,000, while the training time distributional distance decreased from 335.57 to 88.46. These results demonstrate that the model can generate legible words under held-out writer conditioning, but do not directly measure writer-style preservation.

CER measures content legibility by comparing the recognizer's output with
the conditioning word. It does not directly measure preservation of the
conditioning writer's style. Because the two word-generation conditions
are symmetric and differ only in the sampled words, the observed CER
difference may reflect differences in word-level recognition difficulty.
We therefore do not interpret these CER values as a quantitative measure
of writer-style preservation.

\subsubsection{Handwriting-domain distributional evaluation}

For final distributional evaluation, we compute HFID, KID, and
cross-set cosine similarity in the feature space of
TrOCR-base-handwritten~\cite{li2023trocr}. Because no public
Devanagari-specific handwriting encoder is available, these metrics are
encoder-relative and inherit the Latin-script (IAM) prior of TrOCR.
Table~\ref{tab:hfid_metrics} reports the results over held-out real
word images and generated samples.

\begin{table}[t]
\centering
\caption{Handwriting-domain distributional metrics computed in TrOCR
embedding space.}
\label{tab:hfid_metrics}
\resizebox{\columnwidth}{!}{%
\begin{tabular}{lc}
\toprule
Metric & Value \\
\midrule
HFID (95\% CI, bootstrap $n{=}1000$) & 47.09 [46.82, 47.43] \\
KID & $0.234 \pm 0.011$ \\
\midrule
Cosine sim. (real--real) & 0.929 \\
Cosine sim. (gen--gen) & 0.955 \\
Cosine sim. (real--gen) & 0.844 \\
\bottomrule
\end{tabular}
}
\end{table}

The ``real--generated" cosine similarity (0.844) is somewhat lower than
``real--real" (0.929) and ``generated--generated" (0.955), but remains
consistent with substantial overlap in the embedding space. These
metrics are computed over writer-pooled samples and therefore assess
population-level similarity rather than per-writer style preservation.
Embedding-norm analysis is provided in Section 8.6 in Supplementary Material.

\section{Data Validation}
\label{sec:data_validation}
\subsection{Quality Requirements}
\label{subsec:quality_assessment}

We developed an automated image-quality assessment system using CNN classifiers (see Section~5 in Supplementary Material) to identify low-quality pages based on image characteristics. The classifier serves as a reproducible dataset-curation utility rather than a novel quality-assessment method. We use a compact, fully specified architecture that can be readily retrained and audited instead of a pretrained backbone requiring additional validation for handwriting-quality cues. The lack of a pretrained-backbone comparison is a limitation of this curation component and is left for future work.

Image sharpness was quantified using Laplacian variance~\cite{PechPacheco2000}, defined as $\text{Blur Score} = \text{Var}(\nabla^2 I)$, where sharp strokes produce higher variance and blurred strokes produce lower variance.

\subsubsection{Dataset Quality Characterization and Labeling}
Blur scores across all 531 samples ranged from 9.7 to 13,865 ($\mu$=3268, median=4110, $\sigma$=2438), reflecting acquisition variability (camera vs. scanner, lighting, ink, paper). Since blur score alone cannot capture ink density, stroke continuity, or background noise~\cite{Szandala2020}, we use it only for initial labeling and train CNNs on raw pixels for the final quality classification.

\begin{figure}[t]
    \centering
    \includegraphics[scale=0.3]{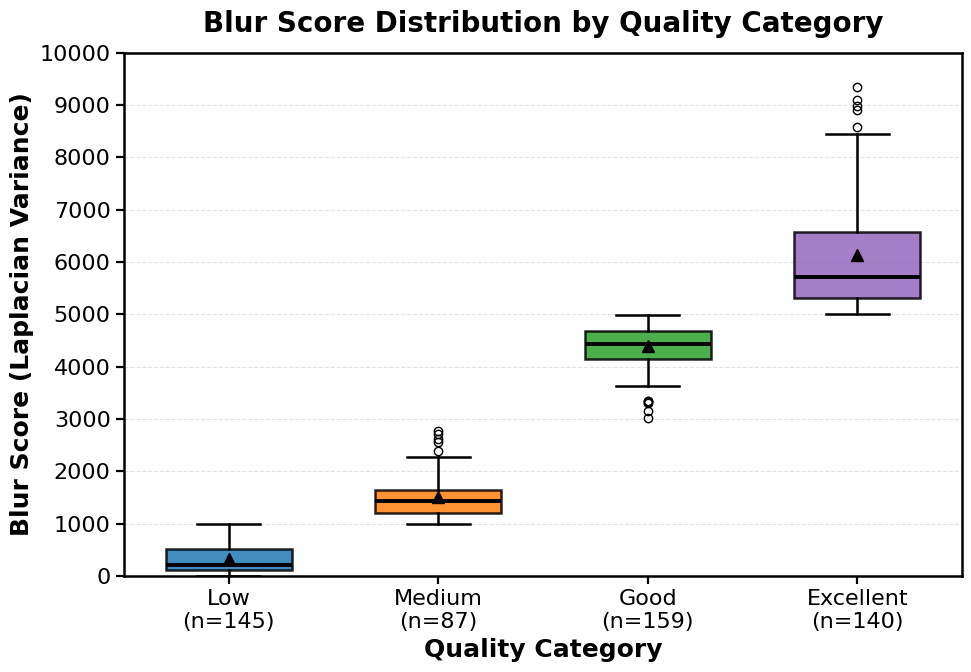}
    \caption{Blur score by quality class}
    \label{fig:blur_boxplot}
\end{figure}

Based on the score distribution, we selected 3000 (43rd percentile) as the initial binary threshold and validated it through manual inspection of 100 boundary-region samples spanning the 2500--3500 range, confirming that 3000 effectively separates degraded images from those with sufficient stroke clarity; this yields 232 (43.7\%) Low-quality and 299 (56.3\%) High-quality images. Four-class thresholds at 1000/3000/5000 were selected analogously (Section 5 in Supplementary Material). The \textit{binary classifier} separates Low-to-Medium ($<3000$) from High ($\geq3000$) and the \textit{four-class classifier} stratifies into Low/Medium/Good/Excellent. Class sizes and per-class statistics are given in Figure~\ref{fig:blur_boxplot}. The dataset was stratified into training (339), validation (85), and test (107) sets. Architecture and training details are in Section 5 in Supplementary Material.

\subsubsection{Results and Discussion}
\paragraph{Classification Performance:}
Table~\ref{tab:quality_overall} summarizes test-set performance (class sizes in Figure~\ref{fig:blur_boxplot}). The binary classifier reached 96.26\% accuracy with balanced precision/recall/F1. The four-class model reached 85.98\%, with the Medium class weakest (F1=0.79), consistent with Medium--Good boundary ambiguity. Validation and test accuracy were closely aligned, suggesting limited overfitting.

\begin{table}[t]
\centering
\caption{Quality classification performance on the held-out test set (107 samples).}
\label{tab:quality_overall}
\resizebox{\columnwidth}{!}{%
\begin{tabular}{llcccc}
\hline
\textbf{Model} & \textbf{Class (threshold)} & \textbf{Test Acc.} & \textbf{Prec.} & \textbf{Rec.} & \textbf{F1} \\
\hline
\multirow{2}{*}{CNN (Binary)}
& Low-Med. ($<3k$) & 96.26\% & 0.94 & 0.99 & 0.96 \\
& High ($\geq3k$) & & 0.99 & 0.93 & 0.96 \\
\hline
\multirow{4}{*}{CNN (Four-class)}
& Low ($<1k$) & 85.98\% & 0.96 & 0.90 & 0.93 \\
& Medium (1--3k) & & 0.75 & 0.83 & 0.79 \\
& Good (3--5k) & & 0.93 & 0.81 & 0.87 \\
& Excellent ($\geq5k$) & & 0.78 & 0.89 & 0.83 \\
\hline
\end{tabular}%
}
\end{table}


\paragraph{Quality-Based Dataset Filtering:}
The binary classifier (confidence $\geq$0.7) retained 288/531 images (54.2\%) as a high-quality core: 157 Good + 131 Excellent, mean blur $>$5000. That only 288 of the 299 images exceeding the 3000 blur threshold were retained indicates the CNN captures quality cues beyond blur alone. The remaining 243 images (mean blur $<1500$) form a degraded subset useful for robustness and restoration studies.

\begin{figure*}[t]
\centering
\begin{tcolorbox}[colframe=black!60,colback=white,boxrule=0.8pt,sharp corners,left=2pt,right=2pt,top=1pt,bottom=1pt,width=\linewidth]
\noindent
\begin{minipage}[b]{0.32\linewidth}
    \centering
    \includegraphics[width=\linewidth]{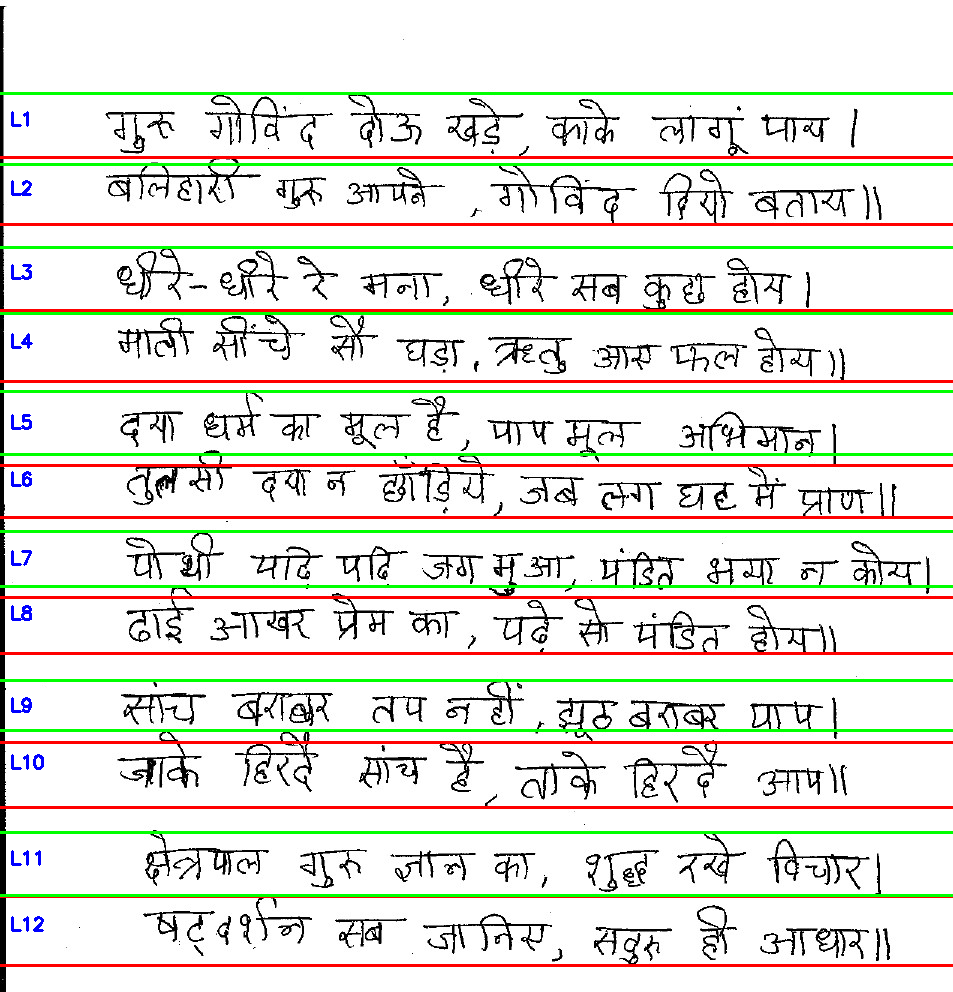}
    \vspace{1mm}
    {\small \textbf{Easy}}
\end{minipage}%
\hfill\vrule width 0.8pt\hfill
\begin{minipage}[b]{0.32\linewidth}
    \centering
    \includegraphics[width=\linewidth]{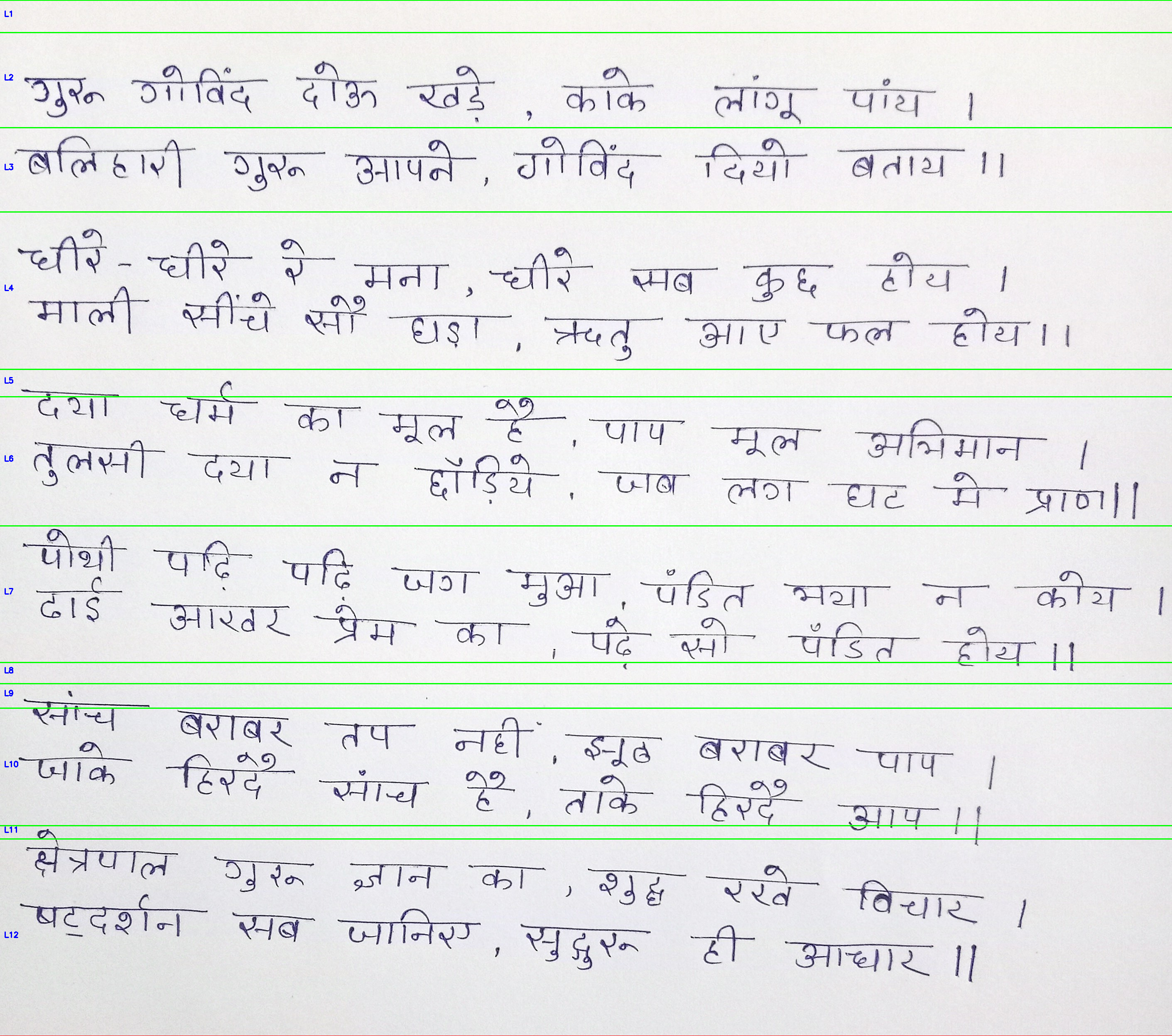}
    \vspace{1mm}
    {\small \textbf{Medium}}
\end{minipage}%
\hfill\vrule width 0.8pt\hfill
\begin{minipage}[b]{0.32\linewidth}
    \centering
    \includegraphics[width=\linewidth]{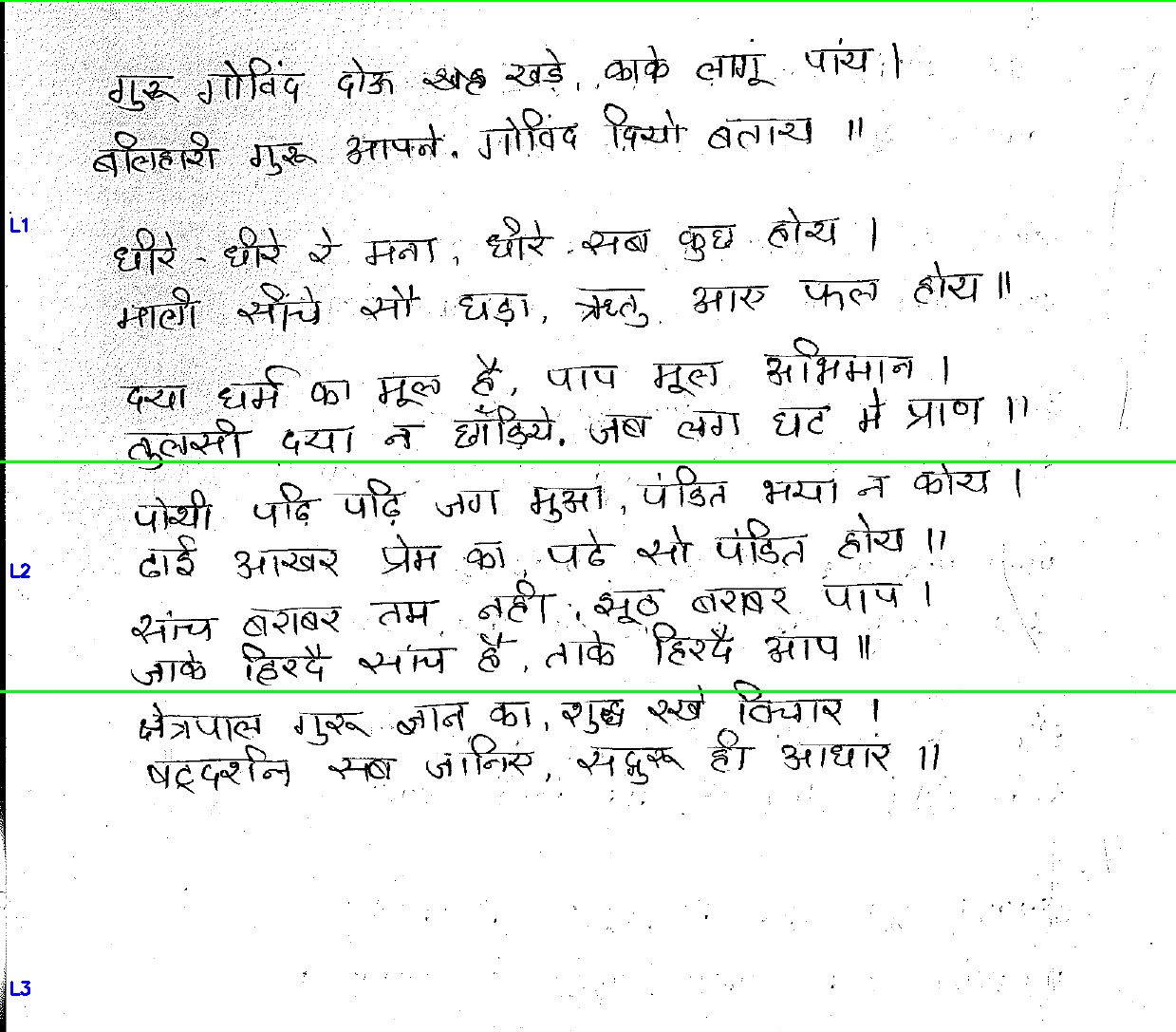}
    \vspace{1mm}
    {\small \textbf{Complex}}
\end{minipage}
\end{tcolorbox}
\caption{Representative pages per segmentation difficulty class. Easy: clear inter-line separation. Medium: moderate spacing variation. Complex: overlap and irregular baselines.}
\label{fig:seg_examples}
\end{figure*}

\subsection{Data Variability}
\label{sec:data_variability}

Quality filtering does not capture structural variability in handwriting, such as inter-line spacing, baseline stability, stroke overlap, and \emph{shirorekha} continuity, which can affect line separability independently of overall image quality~\cite{Romero2015}. We therefore analyze segmentation difficulty across the full dataset to characterize structural variation that may affect downstream handwritten text recognition.

\subsubsection{Experimental Setup}
Each of the 531 raw pages (12 lines each) is processed with a hybrid heuristic segmentation pipeline and difficulty labels are assigned from the resulting line structures (details in Section 7 in Supplementary Material).

\paragraph{Difficulty Annotation Protocol.} Each page receives a composite score $S \in [0,100]$ from line-count accuracy and layout regularity cues. Using $S$ and absolute line-count error, we define \textbf{Easy} ($\leq$1 line error, $S\geq65$), \textbf{Medium} ($\leq$3 line error, $S>45$), and \textbf{Complex} (remaining). Thresholds were chosen empirically. Representative samples appear in Figure~\ref{fig:seg_examples}, with detected boundaries marked by green (upper) and red (lower) lines.



\subsubsection{Results and Discussion}
\label{sec:segmentation_results}

Across the 531 pages, perfect segmentation was achieved for 157 pages (29.57\%). The mean segmentation score was $43.46\pm22.55$ (95\% CI: [41.54, 45.39]), with a median of 42.91. On average, the pipeline detected $11.22\pm2.92$ lines per page, indicating substantial variability in page-level layout and segmentation difficulty.

\textbf{Segmentation difficulty distribution.} Of the 531 pages, 110 (20.7\%) are Easy (score 65.1–91.8), 141 (26.6\%) are Medium (45.9–66.4), and 280 (52.7\%) are Complex (7.6–46.7). Thus, over half of the pages exhibit substantial segmentation difficulty.


Complexity largely arises from intrinsic handwriting features (dense spacing, overlapping strokes, baseline fluctuation, shirorekha-induced continuity) rather than acquisition noise. 55.7\% of pages have $\geq$2 line detection errors (full distribution in Section 7 in Supplementary Material), confirming a non-trivial tail of challenging layouts. These Easy/Medium/Complex annotations provide a practical benchmark for stratified evaluation and motivate adaptive segmentation models.

\subsection{Inter-Rater Validation}
\label{subsec:inter_rater}

Both the quality thresholds and segmentation difficulty labels were independently validated by human judgment. Two blinded annotators labeled random subsets, with inter-rater agreement measured using Cohen's $\kappa$. For \textbf{quality}, the annotators independently labeled 60 boundary-region samples as acceptable or unacceptable, achieving 95\% agreement (57/60; $\kappa=0.89$). For \textbf{segmentation difficulty}, they labeled 60 pages as Easy, Medium, or Complex, achieving 91.7\% agreement (55/60; $\kappa\approx0.84$). Disagreements were concentrated in borderline cases, supporting that both annotation schemes capture visually meaningful distinctions in the collected data.

\section{Dataset availability}

The \emph{DohaScript} dataset is publicly available via an Huggingface\footnote{\url{https://huggingface.co/datasets/SIR-Lab/DohaScript}} repository.


\section{Conclusion}
We presented \emph{DohaScript}, a multi-writer dataset of continuous handwritten Hindi text collected from 531 contributors across 24 states of India, with every writer transcribing the same six traditional \textit{dohas}. This parallel design enables controlled analysis of writer-specific variation while holding linguistic content constant. Our OCR experiments show that continuous handwritten Devanagari remains challenging even for the SOTA systems, while comparison with synthetic handwriting highlights the substantial impact of authentic handwriting variability on recognition performance.

\emph{DohaScript} is also supported by automated quality filtering and page-level segmentation-difficulty annotations, both independently validated by human raters. A word-level GANWriting baseline demonstrates the dataset's utility for writer-conditioned handwriting generation under held-out writer conditioning, while we deliberately do not interpret the current generation results as direct evidence of individual writer-style preservation. Together, these resources provide a controlled, quality-validated benchmark for studying writer variation, recognition, segmentation, and generative modeling in continuous Devanagari handwriting. \emph{DohaScript} is intended to complement, rather than replace, lexically diverse HTR datasets.

\noindent \textbf{Dataset and Code Availability.}
A representative subset of the \emph{DohaScript} dataset is provided in the supplementary material for review. The complete dataset and accompanying code will be publicly released upon acceptance.

\bibliographystyle{ACM-Reference-Format}
\bibliography{main}

\appendix

\section{Dohas}
\label{sec:dohas}

\begin{center}
1.\\

\includegraphics[width=0.3\textwidth]{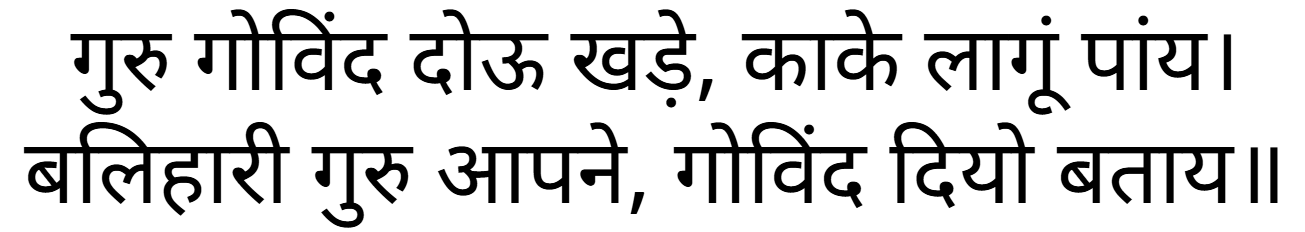}

When both the Guru and God stand before the seeker, whom should one revere first? \\ I bow to the Guru, for it is through the Guru’s guidance that one attains knowledge of God.

2.\\

\includegraphics[width=0.3\textwidth]{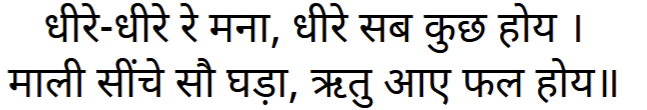}

The mind is urged to remain patient, as everything happens gradually and in its proper time.\\ Even if a gardener waters a plant a hundred times, fruit appears only when the proper season arrives.

3.\\

\includegraphics[width=0.3\textwidth]{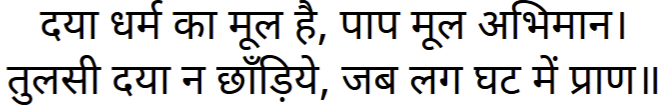}

Compassion is the root of righteousness, and arrogance is the root of sin.\\ Compassion should not be abandoned for as long as life remains in the body.

4.\\

\includegraphics[width=0.3\textwidth]{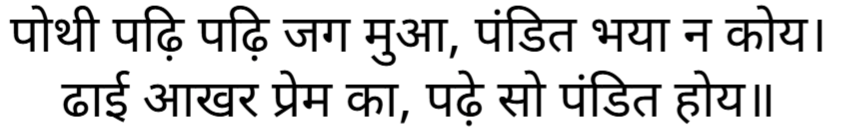}

The world has spent its life reading books, yet no one became truly learned by that alone.\\ One who understands the two and a half letters that form the word “love” (in Hindi) becomes truly learned.

5.\\

\includegraphics[width=0.3\textwidth]{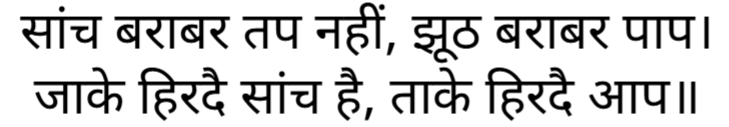}

There is no austerity equal to truth, and no sin equal to falsehood.\\ One whose heart holds truth has God dwelling within their heart.

6.\\
\includegraphics[width=0.3\textwidth]{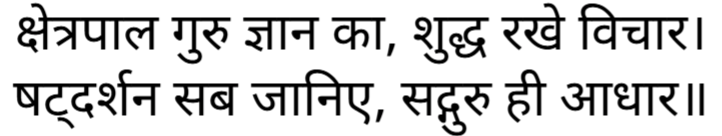}

The Guru acts as the guardian of knowledge and keeps one’s thoughts pure.\\ Even if all six philosophical systems are known, the true Guru alone is the foundation.
\end{center}

\section{Age-wise Participant Distribution}
\label{sec:age_frequency}

\begin{figure}[h]
    \centering
    \includegraphics[width=0.7\linewidth]{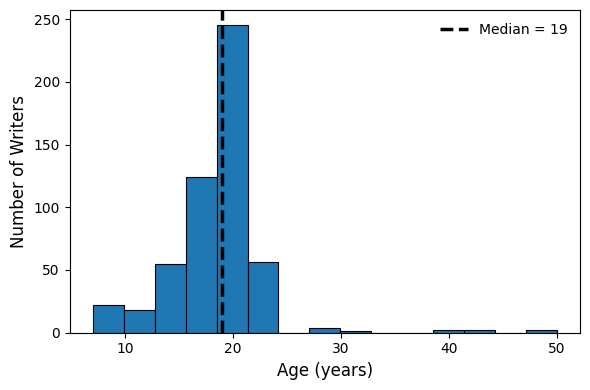}
    \caption{Age distribution of the 531 participants in the dataset. The dashed line indicates the median age. }
    \label{fig:age_distribution}
\end{figure}

 Figure~\ref{fig:age_distribution} and Table ~\ref{tab:age_frequency} shows the age-wise distribution of the participants.

\begin{table}[h]
\centering
\small
\caption{Distribution of participants across age groups.}
\label{tab:age_frequency}
\begin{tabular}{cc|cc} 
\toprule
\textbf{Age} & \textbf{Participants (\#)} & \textbf{Age} & \textbf{Participants (\#)} \\
\midrule
7  & 4  & 19 & 99 \\
8  & 4  & 20 & 76 \\
9  & 14 & 21 & 70 \\
10 & 2  & 22 & 47 \\
11 & 7  & 23 & 6  \\
12 & 9  & 24 & 3  \\
13 & 22 & 28 & 3  \\
14 & 26 & 29 & 1  \\
15 & 7  & 31 & 1  \\
16 & 3  & 40 & 1  \\
17 & 43 & 41 & 1  \\
18 & 78 & 42--50* & 4 \\ 
\bottomrule
\end{tabular}
\end{table}
 
\section{State-wise Participant Distribution}
\label{sec:state_frequency}

\begin{figure}[t]
    \centering
    \includegraphics[width=0.85\columnwidth]{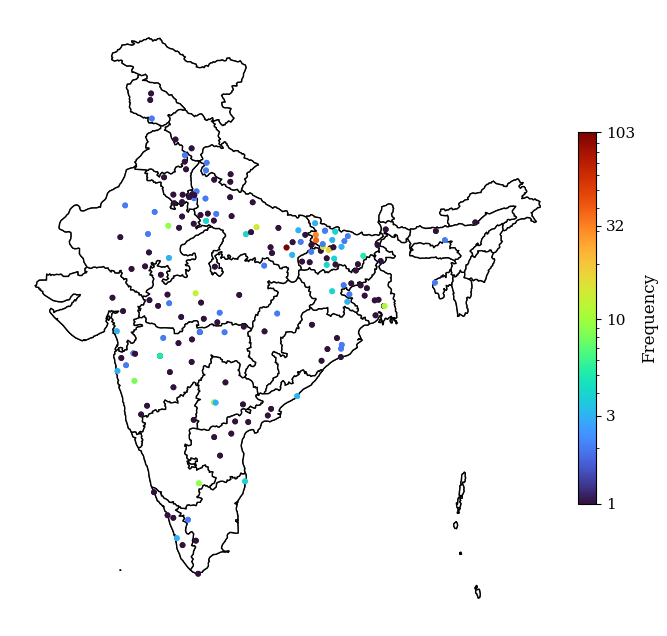}
    \vspace{-0.5em}
    \caption{Frequency distribution of handwriting samples across states of India.}
    \label{fig:state_frequency}
\end{figure}

 Table~\ref{tab:state_frequency} and Figure~\ref{fig:state_frequency} shows the state-wise distribution of the participants.

\begin{table}[t]
\centering
\small
\caption{State-wise distribution of participants.}
\label{tab:state_frequency}
\setlength{\tabcolsep}{4pt}
\begin{tabular}{l c | l c}
\toprule
\textbf{State} & \textbf{Count} & \textbf{State} & \textbf{Count} \\
\midrule
Andhra Pradesh    & 15  & Karnataka         & 9   \\
Arunachal Pradesh & 1   & Kerala            & 7   \\
Assam             & 3   & Madhya Pradesh    & 28  \\
Bihar             & 150 & Maharashtra       & 36  \\
Chandigarh        & 2   & Odisha            & 8   \\
Chhattisgarh      & 3   & Rajasthan         & 25  \\
Delhi             & 3   & Tamil Nadu        & 8   \\
Gujarat           & 5   & Telangana         & 14  \\
Haryana           & 10  & Tripura           & 2   \\
Himachal Pradesh  & 2   & Uttar Pradesh     & 155 \\
Jammu \& Kashmir  & 4   & Uttarakhand       & 7   \\
Jharkhand         & 11  & West Bengal       & 23  \\
\bottomrule
\end{tabular}
\end{table}

\section{Baseline OCR Metrics}
\label{app:metrics}

Performance was evaluated using Character Error Rate (CER) and Word Error Rate (WER), defined as:

\begin{equation}
\mathrm{CER} = \frac{S_c + D_c + I_c}{N_c}
\end{equation}

\begin{equation}
\mathrm{WER} = \frac{S_w + D_w + I_w}{N_w}
\end{equation}

where (S), (D), and (I) denote substitution, deletion, and insertion counts, and (N) is the number of ground-truth units (characters or words).

\section{Quality Requirements Architecture}
\label{app:quality_architecture}

\begin{figure*}[!t]
    \centering
    \includegraphics[width=\textwidth]{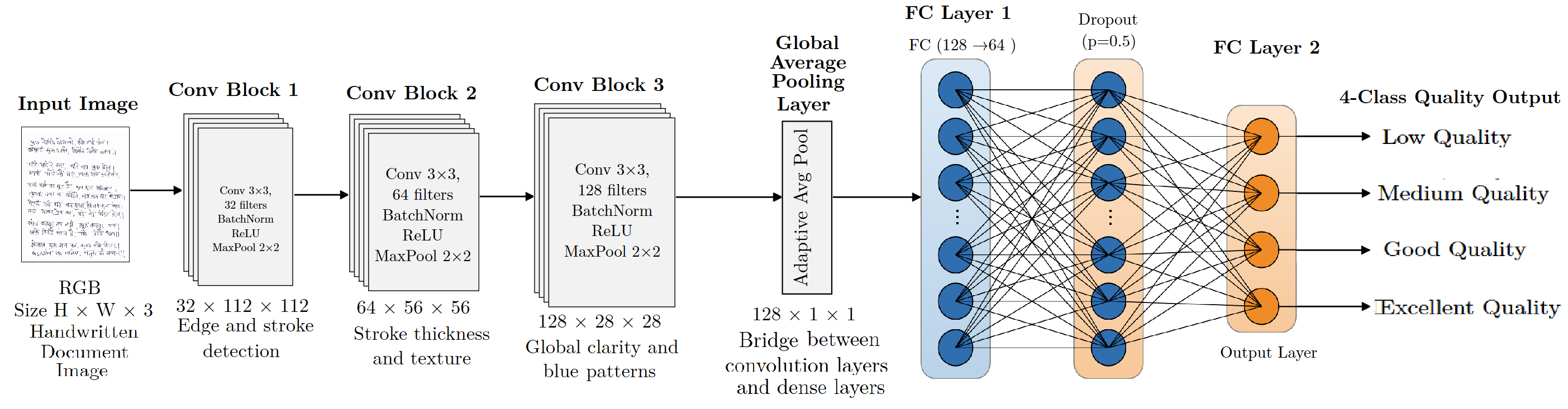}
    \caption{CNN architecture for handwriting quality classification. Three convolutional blocks extract hierarchical features, followed by adaptive pooling and fully connected layers for quality prediction.}
    \label{fig:cnn_arch}
\end{figure*}

The quality classifiers share a common lightweight convolutional backbone (Figure~\ref{fig:cnn_arch}). The input is a resized RGB handwriting-document image ($H \times W \times 3$). Three convolutional blocks extract features at increasing depth and decreasing spatial resolution: Block 1 (32 filters, $3\times3$, BatchNorm, ReLU, MaxPool $2\times2$) targets edge and stroke detection; Block 2 (64 filters, $3\times3$, BatchNorm, ReLU, MaxPool $2\times2$) captures stroke thickness and texture; Block 3 (128 filters, $3\times3$, BatchNorm, ReLU, MaxPool $2\times2$) captures global clarity and blur patterns. An adaptive global average pooling layer bridges the convolutional and dense layers, producing a $128\times1\times1$ feature vector. This is passed through a fully connected layer (128$\rightarrow$64) with dropout ($p=0.5$), followed by an output layer with 2 units (binary classifier) or 4 units (four-class classifier).

Both classifiers were trained in PyTorch 2.0 on CUDA-enabled GPUs using the Adam optimizer and cross-entropy loss, with the dataset stratified into training (339, 63.8\%), validation (85, 16.0\%), and test (107, 20.2\%) splits to ensure balanced class representation and prevent imbalance-induced bias.

Threshold selection for quality labeling was based on iterative analysis of the blur score distribution combined with visual validation. For the binary classifier, a threshold of 3000 (43rd percentile) was validated by manual inspection of 100 boundary-region samples spanning the 2500--3500 range, yielding balanced classes of 232 (43.7\%) low-quality and 299 (56.3\%) high-quality images. For the four-class classifier, thresholds of 1000, 3000, and 5000 were selected based on distributional quartiles and visual assessment, yielding class sizes of 145 (27.3\%) Low, 87 (16.4\%) Medium, 159 (29.9\%) Good, and 140 (26.4\%) Excellent samples, with mean blur scores of 294, 1432, 4269, and 6129 respectively.

\section{OCR Baseline: Implementation Details}
\label{app:ocr_details}

\subsection{Preprocessing}

All experiments use the standardized \emph{DohaScript} images
(Section 3 in main paper). For Tesseract, we applied a unified
preprocessing pipeline consisting of CLAHE contrast enhancement, Otsu
binarization, and morphological denoising. TrOCR receives the original RGB
images directly, consistent with its model design. Sarvam Document
Intelligence also receives the original images without external
preprocessing, as the system incorporates its own image-processing
pipeline.

These differences reflect the intended operating conditions of each
system rather than a controlled preprocessing ablation. Our objective is
to evaluate each system under a practical configuration rather than to
attribute performance differences to a common preprocessing pipeline.

\subsection{Tesseract}

We evaluate four Tesseract~\cite{smith2007overview} configurations:

\begin{enumerate}
    \item \texttt{default}: Hindi language pack with automatic page
    segmentation;
    \item \texttt{OEM 3, PSM 6}: LSTM-based recognition with a uniform
    text-block assumption;
    \item \texttt{PSM 4}: single-column segmentation; and
    \item \texttt{Hindi+English}: bilingual recognition using
    \texttt{hin+eng}.
\end{enumerate}

\subsection{TrOCR}

We evaluate the
\texttt{microsoft/trocr-base-handwritten} checkpoint~\cite{li2023trocr}.
The model was pretrained and fine-tuned on English handwriting and is
applied directly to Devanagari handwriting without Devanagari-specific
pretraining or fine-tuning. Thus, the experiment constitutes a
zero-shot cross-script evaluation. Original RGB images are provided as
input, and predictions are generated using beam search with a beam width
of $k=4$.

\subsection{Sarvam AI}
We evaluate Sarvam Document Intelligence as a contemporary Indic-script
OCR baseline. Images are submitted directly to the Sarvam Vision model
through the Sarvam API using an API key, without external preprocessing. Because the returned predictions may contain Unicode encoding variations, predictions are normalized using Unicode canonicalization before evaluation. We evaluate the minimum CER across the resulting normalization variants to reduce the effect of encoding inconsistencies.

Since the system was released after \emph{DohaScript} data collection, it
serves as a contemporary reference baseline rather than a model
developed specifically for the dataset.

\section{Segmentation Pipeline Details}
\label{app:segmentation_details}
This section provides full implementation details of the heuristic line segmentation strategies and scoring procedure used for dataset-level difficulty annotation.

\subsection{Projection Profiles}
We use horizontal projection profiles as a structural cue for text-line organization~\cite{LikformanSulem2007,Ptak2017Projection}. Given a binarized page image $B(x,y)$, the horizontal projection is defined as:
\begin{equation}
P_h(y) = \sum_x B(x,y),
\end{equation}
which measures foreground ink density along the vertical axis. Distinct peaks typically correspond to individual text lines, whereas compressed spacing or overlapping strokes lead to merged responses and ambiguous boundaries~\cite{Arivazhagan2007LineSeg}.

\subsection{Hybrid Line Segmentation Strategy}
To handle diverse handwriting layouts, we employ a hybrid line segmentation pipeline combining three complementary classical heuristics.

A projection-based method estimates candidate line boundaries by detecting valleys in Gaussian-smoothed ($\sigma=2$) horizontal projection profiles, reducing spurious fluctuations~\cite{LikformanSulem2007,Arivazhagan2007LineSeg}.

In parallel, a contour-based strategy groups text regions using horizontal morphological dilation with a $50\times1$ kernel, followed by contour extraction and bounding-box merging~\cite{Schneider2021Combining}.

Finally, a morphology-driven approach suppresses shirorekha-like horizontal strokes and applies connected component grouping to better isolate Devanagari text lines~\cite{Shinde2014Shirorekha}.

All three strategies are applied independently to each page, and the segmentation whose detected line count is closest to the expected 12 lines is selected.

\subsection{Composite Difficulty Score}
Each page is assigned a composite heuristic score $S \in [0,100]$ based on structural cues extracted from the detected line regions:
\begin{equation}
S = S_{\text{count}} + S_{\text{height}} + S_{\text{spacing}} + S_{\text{coverage}} + S_{\text{straight}},
\end{equation}
where the terms represent line count accuracy, line height uniformity, spacing consistency, coverage ratio, and boundary regularity.

The complete scoring implementation is released with the dataset codebase for reproducibility.

\subsection{Segmentation Error Distribution}

Table ~\ref{tab:error_distribution} shows the distribution of line detection errors.

\begin{table}[h]
\centering
\caption{Distribution of line detection errors.}
\label{tab:error_distribution}
\begin{tabular}{ccc}
\hline
\textbf{Error (lines)} & \textbf{Count} & \textbf{Percentage} \\
\hline
0 & 157 & 29.6\% \\
1 & 138 & 26.0\% \\
2 & 69 & 13.0\% \\
3 & 64 & 12.1\% \\
4 & 28 & 5.3\% \\
5 & 28 & 5.3\% \\
6 & 10 & 1.9\% \\
7 & 18 & 3.4\% \\
8 & 9 & 1.7\% \\
9 & 7 & 1.3\% \\
10 & 2 & 0.4\% \\
11 & 1 & 0.2\% \\
\hline
\end{tabular}
\end{table}

\section{Handwriting Generation Baseline: Additional Details}
\label{app:ganwriting_details}

This section of supplementary material provides the model adaptation and training details for the handwriting generation baseline as discussed in Section 5 of main paper.

\subsection{Word-Level Scope}
\label{app:word_level_scope}

\emph{DohaScript} is collected and released at the page level, and the OCR experiments operate on continuous handwritten text. The generation baseline, however, trains and evaluates at the word level. This follows the word-conditioned design of GANWriting~\cite{kang2020ganwriting} and provides a larger and more uniform set of training samples through the line- and word-segmentation pipeline described in Section 6.2 of main paper.

Adapting GANWriting to full pages would require a different generative architecture capable of handling variable-length text, line layout, and page-level spatial structure. We therefore use the word-level setting as a proof-of-utility baseline rather than introducing a new page-level generator. Page-level handwriting generation is left for future work.

Word-level generation also provides a controlled setting for evaluating writer-conditioned generation: a generated word can be conditioned on a writer reference while its lexical content is independently specified. The experiment therefore complements the page-level OCR evaluation by testing whether the dataset can support generative modeling of handwriting style.

\subsection{Train/Test Split}
\label{app:train_test_split}

The split used for this baseline is at the writer level, not the word level. All writer IDs present in the word-level ground truth produced by the segmentation pipeline (Section 6.2 of main paper) are shuffled with a fixed seed and partitioned training set of 85\% and test set 15\% into disjoint train and test writer sets. No writer appears in both splits. Test-set writers are therefore genuinely unseen during training, and the reported CER values reflect generation conditioned on writer styles that were not observed during training.

The word vocabulary is not split. Every writer, including both train and test writers, transcribes the same fixed six-doha, 89-word corpus. The candidate words used for both the directly generated and swapped conditions are drawn from this shared vocabulary. Consequently, the swapped-word condition does not represent out-of-vocabulary generalization. Instead, it evaluates generation of a different word from the same fully-seen vocabulary while conditioning on a held-out writer's style.

\subsection{Model and Adaptation}

We adapt GANWriting~\cite{kang2020ganwriting} to
\emph{DohaScript}, training on word-level images paired with their Unicode transcriptions. The adaptation replaces the original Latin-script vocabulary with Devanagari characters and modifies text rendering to support the \emph{shirorekha}, vowel diacritics, and conjunct characters.
The input representation and image dimensions are also adjusted for the horizontal extent of Hindi words. The generator, style encoder, discriminator, and auxiliary recognizer otherwise follow the original GANWriting framework.

The training and evaluation writer partitions are disjoint. Thus, the writer references used for evaluation correspond to writers whose handwriting images were not used for model training. The lexical vocabulary, however, is fixed and fully observed during training; the generation experiment therefore evaluates held-out writer conditioning rather than novel lexical content.


\subsection{Training Dynamics}
\label{app:training_dynamics}
The model was trained for 50{,}000 adversarial steps. CER and a
training-time distributional-distance proxy were periodically computed on a fixed held-out evaluation set. The proxy uses Inception-v3 features and is included only as a lightweight measure of training progress; it is not used as the primary distributional evaluation reported in the main paper.

During the initial training phase (epochs 0--400), CER decreased from 100\% to approximately 18\%, while the training-time distance decreased from 335.6 to approximately 130. During subsequent training, CER fluctuated between approximately 15--45\%, while the distance continued to decrease, reaching approximately 77--80 in later training. Training remained stable throughout the 50{,}000 adversarial steps without observed discriminator or generator collapse.

\subsection{Handwriting-Domain Distributional Evaluation}
\label{app:hfid_details}

For final distributional evaluation, we compute HFID, KID, and cosine similarity using embeddings extracted from \texttt{TrOCR-base-handwritten}~\cite{li2023trocr}. HFID and KID compare the distributions of real and generated word-image embeddings, while cosine similarity is computed between pairs drawn from the real and
generated sets.

Because the encoder was trained primarily on Latin-script handwriting, these metrics should be interpreted as encoder-relative measures rather than script-independent measures of handwriting similarity. In particular, the absence of a public Devanagari-specific handwriting encoder limits the extent to which these values can be interpreted as absolute measures of Devanagari handwriting quality.

The evaluation uses a large held-out pool of real word images and a
larger set of generated samples from the trained model. HFID confidence intervals are obtained using 1,000 bootstrap resamples. KID is reported with its corresponding sampling variation.

\subsection{Embedding Norm Analysis}
\label{app:embedding_norms}

\begin{figure}[h]
    \centering
    \includegraphics[width=0.9\linewidth]{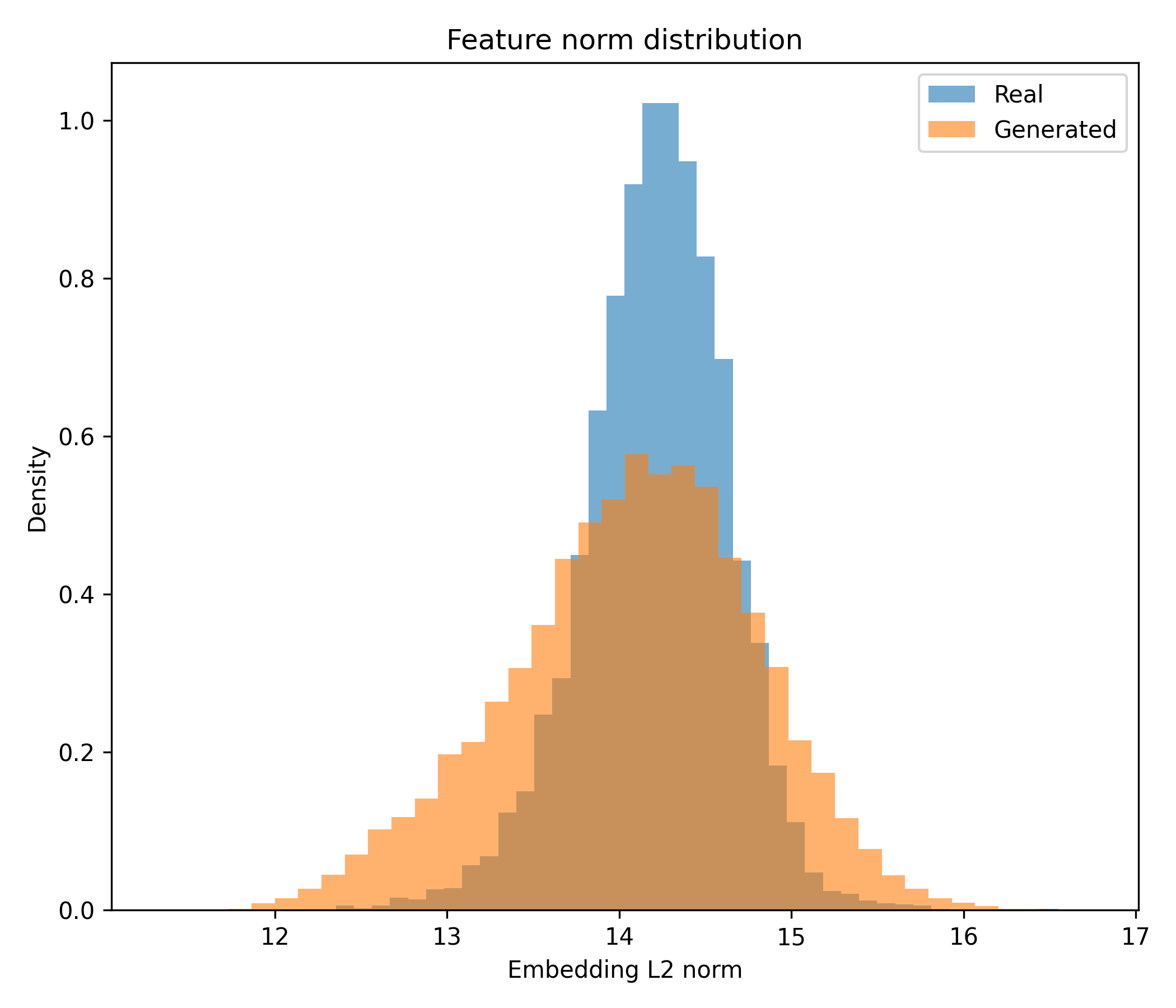}
    \caption{Distribution of embedding L2 norms for real and generated word images in TrOCR feature space.}
    \label{fig:feature_norms}
\end{figure}

To examine whether the real--generated distributional differences reported in Section 5.1 of the main paper (Table 3) could be explained primarily by a systematic difference in embedding magnitude, we compare the L2 norms of real and generated word-image embeddings in TrOCR feature space. As shown in Figure~\ref{fig:feature_norms}, the two distributions overlap substantially and have similar central values, although generated samples exhibit a somewhat wider spread.

This analysis suggests that the distributional gap measured by HFID and cosine similarity is not primarily attributable to a systematic mismatch in embedding magnitude. The observed differences are therefore more consistent with variation in embedding direction and other aspects of the feature distribution rather than a simple difference in feature scale.

\subsection{Interpretation and Limitations}

The generation experiment is intended as a proof-of-utility baseline for \emph{DohaScript}, rather than as a state-of-the-art handwriting generation study. CER evaluates content legibility and does not measure whether a generated sample preserves the conditioning writer's style.
Similarly, HFID, KID, and cosine similarity are computed over
writer-pooled samples and therefore characterize population-level
distributional similarity rather than per-writer style preservation.

We do not report writer-classification accuracy or another independent per-writer style metric in this study. The auxiliary writer classifier used during GANWriting training functions as a training objective and is not treated as an evaluation metric. Consequently, the current baseline demonstrates the feasibility of writer-conditioned generation on \emph{DohaScript}, but does not establish quantitative writer-style preservation or content--style disentanglement.

\end{document}